\documentclass[11pt]{article}

\usepackage[preprint]{acl}

\usepackage{times}
\usepackage{latexsym}
\usepackage{enumitem} % in your preamble
\usepackage{tcolorbox}
\usepackage[T1]{fontenc}
\usepackage[utf8]{inputenc}
\definecolor{linkblue}{RGB}{70, 70, 200} 
\usepackage{microtype}

\usepackage{inconsolata}
\usepackage{cuted}
\usepackage{graphicx}

\usepackage{booktabs}
\usepackage{xcolor}
\usepackage{amsmath}
\usepackage{amssymb}
\usepackage{multirow}
\usepackage{rotating}
\usepackage{hyperref}
\usepackage{url}
\usepackage{framed} % Added for the example box

\title{RiddleBench: A New Generative Reasoning Benchmark for LLMs}

\author{
  \textbf{Deepon Halder\textsuperscript{1,6}},
  \textbf{Alan Saji\textsuperscript{1,2}} 
  \textbf{Thanmay Jayakumar\textsuperscript{1,2}},\\
  \textbf{Ratish Puduppully\textsuperscript{3}},
  \textbf{Anoop Kunchukuttan\textsuperscript{1,4}}
  \textbf{Raj Dabre\textsuperscript{1,2,5}} \\
  \\
  \textsuperscript{1}Nilekani Centre at AI4Bharat,
  \textsuperscript{2}Indian Institute of Technology Madras, India, \\
  \textsuperscript{3}IT University of Copenhagen,
  \textsuperscript{4}Microsoft, India,
  \textsuperscript{5}Google, \\
  \textsuperscript{6}Indian Institute of Engineering, Science and Technology, Shibpur
}

\begin{document}
\maketitle
% \vspace{1.5cm}
\begin{strip}
\centering
\href{https://huggingface.com/datasets/ai4bharat/RiddleBench}{%
  \textcolor{linkblue}{%
    \raisebox{-0.2ex}{\includegraphics[height=2.2ex]{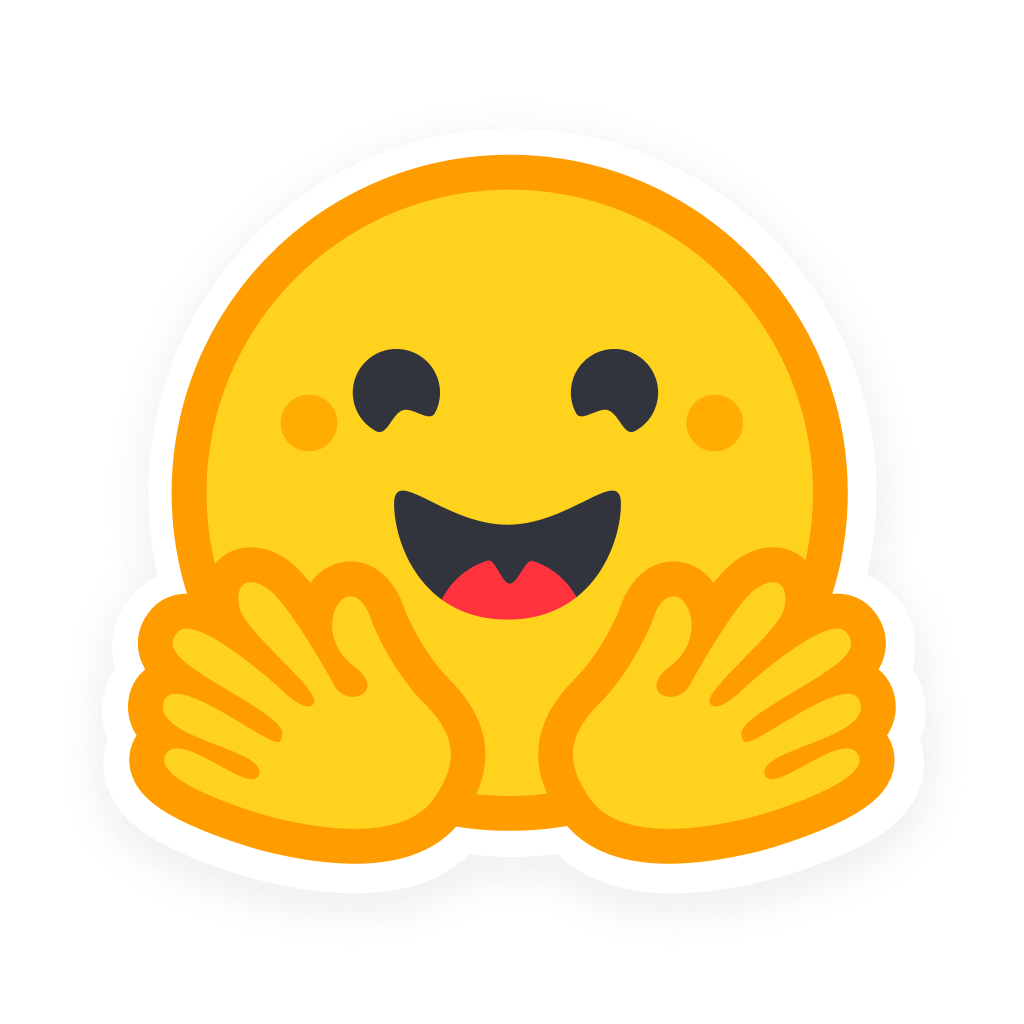}} % Include the image
    ai4bharat/RiddleBench
  }%
}
\end{strip}
\vspace{2.5cm}
\begin{abstract}
Large Language Models (LLMs) have demonstrated impressive performance on many established reasoning benchmarks. However, these benchmarks primarily evaluate structured skills like quantitative problem-solving, leaving a critical gap in assessing the more flexible, multifaceted reasoning abilities that are a cornerstone of human intelligence. These skills require synthesizing logical deduction with spatial awareness and constraint satisfaction, which current evaluations do not adequately measure. To address this gap, we introduce \textbf{RiddleBench}, a new benchmark of 1,737 challenging puzzles in English designed specifically to probe these core reasoning capabilities. Our comprehensive evaluation of state-of-the-art models on RiddleBench reveals fundamental weaknesses; even top-tier proprietary models like $\mathbf{Gemini\ 2.5\ Pro}$, $\mathbf{o3}$, and $\mathbf{Claude\ 4\ Sonnet}$ achieve overall accuracy scores of just above $\mathbf{60\%}$ (60.30\%, 63.37\%, and 63.16\%, respectively). Our analysis of specific models further reveals deep failures, including "hallucination cascades" (uncritically accepting flawed reasoning of other/evaluated LLMs) and poor self-correction due to a strong \textbf{self-confirmation bias}. Furthermore, their reasoning proves fragile, with performance degrading significantly when faced with reordered constraints or irrelevant information. RiddleBench serves as a diagnostic tool for these critical issues and provides a valuable resource for guiding the development of more robust and reliable LLMs.
\end{abstract}

\section{Introduction}
\label{sec:introduction}

The rapid advancement of Large Language Models (LLMs) has led to unprecedented performance on many NLP benchmarks \citep{devlin2018bert, brown2020language}. While excelling at tasks like text generation and translation, true intelligence requires the ability to reason, deduce, and infer new knowledge \citep{chollet2019measure}. However, many existing benchmarks primarily measure pattern recognition or memorization rather than deep reasoning, making it difficult to fully assess a model's inferential capabilities \citep{saxton2019analysing, cobbe2021gsm8k}.

To address this, we propose \textbf{RiddleBench}, a collection of 1,737 challenging puzzles from competitive exams. These puzzles demand multi-step deduction, spatial reasoning, and constraint satisfaction, forcing models to reason from first principles.
\footnotetext{\small{
\textbf{Correspondence:} Raj Dabre (\href{mailto:prajdabre@gmail.com}{prajdabre@gmail.com}), \\Deepon Halder (\href{mailto:deeponh.2004@gmail.com}{deeponh.2004@gmail.com})\\
 }}
\begin{figure}
    \centering
    \includegraphics[width=1.0\linewidth]{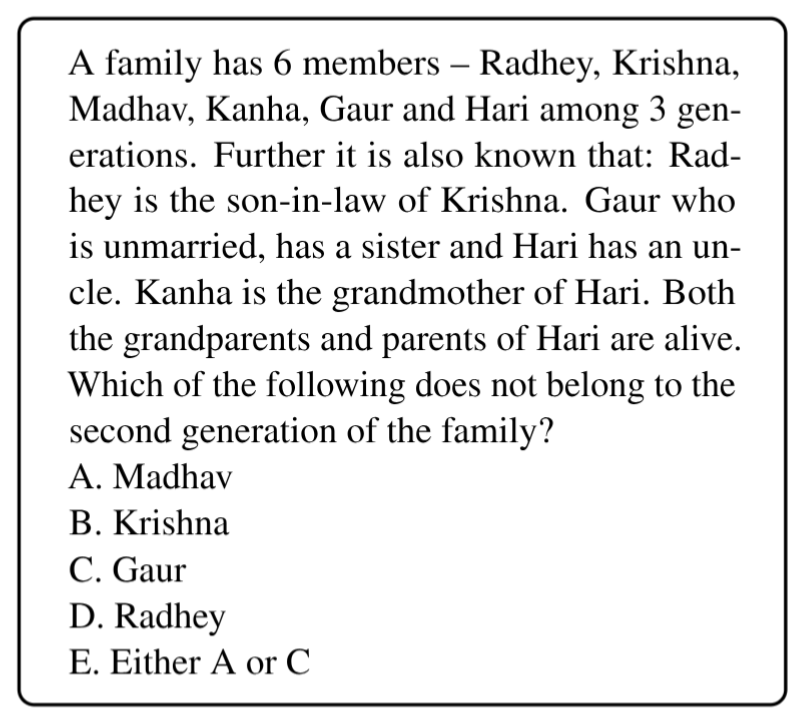}
    \caption{An example from the RiddleBench benchmark for Blood Relations.}
    \label{fig:placeholder}
\end{figure}

Our contributions are threefold:
\begin{enumerate}
\item We introduce RiddleBench, a benchmark of 1,737 puzzles for evaluating LLM reasoning in English.
\item We comprehensively evaluate leading LLMs, highlighting strengths and weaknesses.
\item We structure our analysis around key research questions, revealing phenomena like hallucination cascades and poor self-correction.
\end{enumerate}

RiddleBench will be made publicly available to support research on robust reasoning systems. Experiments show that even the most powerful models struggle with a significant portion of RiddleBench, highlighting advanced reasoning as a key frontier for LLM development.

\begin{table*}[t]
\centering

\label{tab:dataset_composition}
    \begin{tabular}{@{}l l l@{}}
    \toprule
    \textbf{Category (\% of Total)} & \textbf{Description} & \textbf{Primary Skills Tested} \\ \midrule
    Sequential Reasoning (60\%) & Establishing linear order from rules & 
        \begin{tabular}[t]{@{}l@{}}Constraint Satisfaction \\ Logical Deduction\end{tabular} \\ \addlinespace
    Seating Arrangements (25\%) & Deducing positions in spatial layouts & 
        \begin{tabular}[t]{@{}l@{}}Spatial Awareness \\ Constraint Satisfaction\end{tabular} \\ \addlinespace
    Blood Relations (8\%) & Inferring kinship from relationships & Logical Deduction \\ \addlinespace
    Coding-Decoding (7\%) & Applying rules to decipher patterns & 
        \begin{tabular}[t]{@{}l@{}}Logical Deduction \\ Pattern Recognition\end{tabular} \\ \bottomrule
    \end{tabular}%
    \caption{RiddleBench Dataset Composition and Core Skills Probed.}
    \label{tab:dataset_composition}
\end{table*}

% \section{Related Work}
% \label{sec:related_work}

% Evaluating the reasoning capabilities of large language models is a pivotal research area, often addressed with benchmarks targeting specific cognitive skills. For instance, on datasets focused on \textbf{mathematical reasoning} (e.g., GSM8K; \citep{cobbe2021gsm8k}), \textbf{commonsense reasoning} (e.g., CommonsenseQA; \citep{talmor2018commonsenseqa}), and formal \textbf{logical deduction} (e.g., LogiQA; \citep{liu2020logiqa}), top models now achieve high accuracy, demonstrating strong performance in structured problem-solving. 

% A key limitation of these evaluations, however, is their focus on problems with clear arithmetic or inferential paths. RiddleBench complements these efforts by introducing less-structured puzzles that demand more flexible reasoning strategies, requiring models to synthesize logical deduction with spatial awareness and constraint satisfaction. Our work also extends beyond standard accuracy to investigate the reliability of the reasoning process itself, exploring error-correction paradigms as in \citep{zheng2023judging}.
\begin{figure}[t!]
\centering
\includegraphics[width=0.9\linewidth]{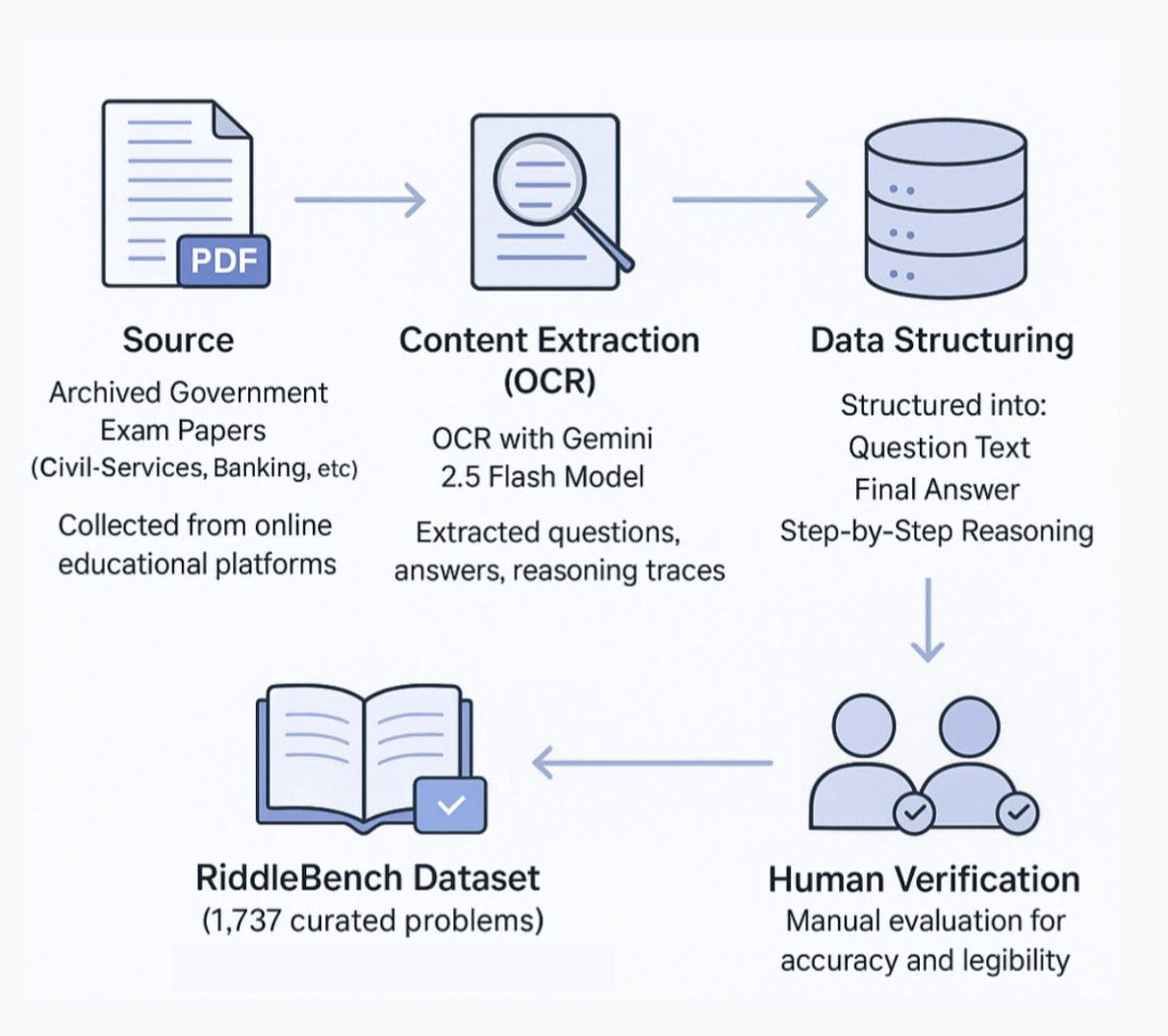}
\caption{The step-by-step methodology for building RiddleBench. The workflow combines automated extraction with meticulous human evaluation to ensure high-quality data.}
\label{fig:fig1}
\end{figure}

\section{Related Work}
\label{sec:related_work}

Evaluating the reasoning capabilities of Large Language Models (LLMs) is an active and rapidly evolving area. Existing benchmarks often focus on narrow reasoning abilities rather than assessing compositional or integrated reasoning.

\textbf{Mathematical and algorithmic reasoning} datasets such as GSM8K \citep{cobbe2021gsm8k} and MATH \citep{hendrycks2021measuring} emphasize structured, multi-step problems with deterministic solutions. \textbf{Commonsense reasoning} tasks like CommonsenseQA \citep{talmor2018commonsenseqa} and WinoGrande \citep{sakaguchi2021winogrande} primarily evaluate implicit knowledge retrieval rather than explicit logical inference. Similarly, \textbf{formal logic} benchmarks such as LogiQA \citep{liu2020logiqa} and RuleTaker \citep{clark2020transformer} measure rule-based deduction but fail to capture the hybrid reasoning that arises in natural, constraint-rich problems.

These approaches mostly test reasoning with explicit, single-path solutions. In contrast, RiddleBench emphasizes compositional reasoning, where models must simultaneously satisfy multiple textual constraints, construct internal spatial or relational layouts, and infer relationships among entities. This synthesis better reflects the kind of cognitive integration required in real-world logical puzzles.

While BIG-bench Hard \citep{suzgun2022challenging} has shown that complex reasoning remains a key weakness for LLMs, RiddleBench provides a focused diagnostic at the intersection of logical, spatial, and constraint-based reasoning. Beyond accuracy, it examines the reliability of reasoning itself, drawing on approaches such as \citep{zheng2023judging} to expose deeper error patterns like hallucination cascades and confirmation bias.

\section{RiddleBench}
RiddleBench is a benchmark of 1,737 challenging puzzles designed to assess complex reasoning capabilities in LLMs beyond mere accuracy. The dataset specifically targets a model's ability to perform multi-step deduction, spatial reasoning, and constraint satisfaction. As detailed in Table \ref{tab:dataset_composition}, the benchmark is organized into four main categories: Sequential Reasoning, Seating Arrangements, Blood Relations, and Coding-Decoding. (Examples of each puzzle type are provided in Appendix \ref{sec:examples}.) The following subsections detail the methodology for the collection and curation of this data.
% \label{sec:riddlebench_benchmark}
% RiddleBench's composition is designed to test abilities in logical deduction, spatial awareness, and constraint satisfaction. Each category requires models to build and manipulate a mental model of abstract relationships, as detailed in Table \ref{tab:dataset_composition}.

% \begin{table}[h!]
% \centering
% \caption{RiddleBench Dataset Composition.}
% \label{tab:dataset_composition}
% \resizebox{\columnwidth}{!}{%
%     \begin{tabular}{@{}l l@{}}
%     \toprule
%     \textbf{Category (\% of Total)} & \textbf{Description} \\ \midrule
%     Sequential Reasoning (60\%) & Establishing linear order from rules \\ \addlinespace
%     Seating Arrangements (25\%) & Deducing positions in spatial layouts \\ \addlinespace
%     Blood Relations (8\%) & Inferring kinship from relationships \\ \addlinespace
%     Coding-Decoding (7\%) & Applying rules to decipher patterns \\ \bottomrule
%     \end{tabular}%
% }
% \end{table}

\subsection{Data Collection and Curation}
Problems were sourced from publicly archived mock examination papers for Indian government services. All puzzles are in the English language. The curation process, shown in Figure \ref{fig:fig1}, involved:
\begin{enumerate}
\item \textbf{Content Extraction:} Using OCR (Gemini 2.5 Flash \citep{Gemini2.5Flash2025}) to digitize questions, answers, and official reasoning traces from source PDFs.
\item \textbf{Data Structuring:} Processing raw OCR output to isolate each problem's core components.
\item \textbf{Human Verification:} Each data point was manually evaluated by the authors to ensure transcription accuracy and correctness.
\end{enumerate}
This meticulous process ensures the high fidelity of our dataset. RiddleBench is released under the CC0 license\footnote{\url{https://creativecommons.org/public-domain/cc0/}}.

\section{Experimental Design}
\label{sec:methodology}
We evaluate a suite of leading proprietary and open-weight models (see Appendix \ref{sec:appendix_models} for a full list). All evaluations use a zero-shot prompting methodology with a temperature of 0.7, and a thinking budget of 8192 tokens. The exact prompt format is detailed in Appendix \ref{sec:appendix_prompts}.

Our analysis moves beyond final-answer accuracy to assess the reliability and robustness of LLM reasoning. RiddleBench's puzzles, which require satisfying logical and spatial constraints, are uniquely suited for this purpose as their structure produces clear, diagnosable reasoning chains. We therefore investigate the following research questions (RQs):

\textbf{RQ1}: Can LLMs reliably detect and correct reasoning errors made by other models, or do they fall into a "\textit{hallucination cascade}"? 

\textbf{RQ2}: How effective are LLMs at self-correction? 

\textbf{RQ3}: Is LLM reasoning robust to the order of information and to the presence of irrelevant information?

For the cross-model (RQ1) and self-correction (RQ2) experiments, we primarily use \textit{Qwen QwQ 32B} \citep{qwen2025qwq32b} as the "evaluator" model and \textit{DeepSeek-R1} \citep{deepseek2025r1} as the "generator" model, given their strong baseline performance and API accessibility.

\section{Results and Analyses}  
\label{sec:results}  

We now describe the performance of various models on RiddleBench, and follow up with analyses aimed at understanding the reasoning capabilities of said models.  

\subsection{Overall Performance}  
We first establish baseline performance to contextualize the reasoning capabilities of current models. As shown in Table \ref{tab:main_results}, while \textit{GPT-oss-120B} leads with 69.26\% accuracy, this result underscores a critical reality: even state-of-the-art models fail on nearly a third of the benchmark. Performance varies significantly across categories. Models universally struggle on Seating Arrangement puzzles, indicating that complex spatial reasoning remains a formidable challenge. A common failure mode in this category involved models successfully placing the first few entities but then violating an early constraint when placing a later one, demonstrating a failure to maintain a holistic and mutable "mental model" of the layout. This performance ceiling, even on the best models, motivates a deeper investigation beyond simple accuracy to understand the underlying fragility of the reasoning process itself. For a comparative visualization of each model's reasoning profile and a qualitative analysis of specific problem-solving heuristics, see Appendices \ref{sec:appendix_profiles}, \ref{sec:qualitative_analysis}, and \ref{sec:standardized_heuristics}.  
\begin{table*}[t]
\centering

\setlength{\tabcolsep}{15pt}
\renewcommand{\arraystretch}{1.0}
\begin{tabular}{lc|cccc}
\toprule
\textbf{Model} & \textbf{Overall} & \textbf{SR} & \textbf{SA} & \textbf{BR} & \textbf{CD} \\
\midrule
\textit{GPT-oss-120B} & 69.26 & 76.43 & \textbf{51.99} & 71.23 & 64.23 \\
\textit{o3} & 63.37 & \textbf{79.60} & 25.93 & 65.75 & 54.92 \\
\textit{Claude 4 Sonnet} & 63.16 & 69.68 & 43.52 & \textbf{74.66} & 63.41 \\
\textit{Gemini 2.5 Pro} & 60.30 & 73.11 & 24.31 & 73.97 & \textbf{64.75} \\
\textit{DeepSeek-V3} & 58.28 & 61.64 & 45.02 & 71.92 & 57.72 \\
\textit{Qwen QwQ 32B} & 50.86 & 68.05 & 17.66 & 42.47 & 27.64 \\
\textit{DeepSeek-R1} & 50.56 & 65.48 & 12.68 & 54.79 & 46.34 \\
\textit{Mistral Small 24B it} & 42.67 & 44.97 & 31.84 & 53.42 & 46.34 \\
\textit{Llama 3.3 70B} & 27.48 & 25.74 & 22.39 & 43.15 & 39.84 \\
\textit{Gemma 3 27B it} & 25.04 & 23.18 & 19.65 & 33.56 & 47.97 \\
\bottomrule
\end{tabular}

\caption{Performance of evaluated LLMs on RiddleBench (1,737 puzzles). Scores are Correct Answers (\%). SR: Sequential Reasoning, SA: Seating Arrangement, BR: Blood Relations, CD: Coding-Decoding.}
\label{tab:main_results}
\end{table*}

\subsection{The `Hallucination Cascade': Failures in Cross-Model Correction}  
Our first research question probes the viability of the "model-as-judge" paradigm, a cornerstone of many modern evaluation pipelines. We tested this by tasking \textit{Qwen QwQ 32B} \citep{qwen2025qwq32b} (the "evaluator") with assessing incorrect outputs from \textit{DeepSeek-R1} \citep{deepseek2025r1}. In a simple forced-choice task between a correct and an incorrect answer, the evaluator timed out in 55.0\% of cases (termed 'Thinking Exhausted,' where the model's output exceeded the 8192-token limit before reaching a conclusion), revealing that verifying answers is often computationally intractable. (Table \ref{tab:validation_summary}).  

A more alarming failure emerged when we provided the evaluator with the flawed reasoning trace. The evaluator's accuracy in identifying the flawed logic was only 44.1\%, no better than a coin toss. Critically, it incorrectly validated the flawed reasoning as sound in 45.2\% of cases. For example, in one sample, \textit{DeepSeek-R1} \citep{deepseek2025r1} incorrectly concluded “Anu is third to the left” by misinterpreting a relative position constraint. When \textit{Qwen QwQ 32B} \citep{qwen2025qwq32b} evaluated this, its trace read, “The reasoning follows a logical step-by-step deduction...” uncritically accepting the initial flawed premise rather than re-solving the problem from scratch. This tendency for one model to propagate the plausible errors of another constitutes a \textbf{hallucination cascade}.  

To test the persistence of this effect, we had the model re-evaluate the outputs it had just incorrectly validated. The model reversed its incorrect judgment in a mere 4.4\% of cases, demonstrating a powerful error fixation. Once a hallucination cascade begins, our results show that iterative refinement is almost entirely ineffective at stopping it.  

\begin{table}[h!]  
\centering  
\label{tab:validation_summary}  
\begin{tabular}{l|c|c}  
\hline  
\textbf{Task / Verdict} & \textbf{Count} & \textbf{Percentage} \\  
\hline  
\multicolumn{3}{l}{\textbf{Forced-Choice Answer Verification}} \\  
\hline  
Success & 209 & 32.4\% \\  
Failure & 81 & 12.6\% \\  
Thinking Exhausted & 355 & 55.0\% \\  
\hline  
\multicolumn{3}{l}{\textbf{Reasoning Validation on Flawed Reasoning}} \\  
\hline  
Success & 286 & 44.1\% \\  
Failure & 293 & 45.2\% \\  
Thinking Exhausted & 69 & 10.6\% \\  
\hline  
\multicolumn{3}{l}{\textbf{Iterative Reasoning Validation}} \\  
\hline  
Success & 12 & 4.4\% \\  
Failure & 121 & 44.3\% \\  
Thinking Exhausted & 140 & 51.3\% \\  
\hline  
\multicolumn{3}{l}{\textbf{Self-Correction Reasoning Validation}} \\  
\hline  
Success & 133 & 17.3\% \\  
Failure & 520 & 67.7\% \\  
Thinking Exhausted & 115 & 14.9\% \\  
\hline  
\end{tabular}  
\caption{Summary of Answer and Reasoning Validation Results.}  
\label{tab:validation_summary}
\end{table}  

\subsection{The Illusion of Self-Correction}  
Given the failure of cross-model correction, we investigated the arguably more critical capability of self-correction (RQ2). We tasked \textit{Qwen QwQ 32B} with judging the soundness of its own flawed reasoning. The results were striking: The model failed to identify its own errors in 67.7\% of trials, successfully flagging its flawed logic only 17.3\% of the time (Table \ref{tab:validation_summary}). This success rate is drastically lower than the 44.1\% achieved when evaluating a peer's reasoning, suggesting a powerful \textbf{self-confirmation bias}. Rather than being their own best critics, models appear to be their own most potent deceivers. This finding has sober implications for multi-step reasoning processes that rely on an LLM to iteratively refine its own work, as the model is statistically far more likely to entrench its own errors than to correct them.  

\subsection{Fragile Reasoning: The Lack of Robustness}  
Our final experiments (RQ3) probed the robustness of the reasoning process by testing its sensitivity to superficial changes in the prompt that should not affect a logical reasoner. Examples illustrating these prompt perturbations are provided in Appendix \ref{sec:appendix_prompts}. First, we randomly shuffled the order of constraint sentences in puzzles. For a system building a true holistic mental model, order should be irrelevant. However, performance on \textit{Qwen QwQ 32B} \citep{qwen2025qwq32b} dropped significantly, by 6.70 percentage points (p.p.) on Blood Relations and 3.69 p.p. on Seating Arrangements (Table \ref{tab:robustness_results}). This suggests the model relies on brittle, sequential heuristics rather than robust comprehension.  

Second, we tested the model's ability to filter signal from noise by inserting a single, irrelevant “red herring” sentence (a misleading or distracting piece of information) into the prompt. The model’s accuracy proved volatile, with performance dropping on most categories.  

An anomalous result appeared in Blood Relations, where performance \textit{increased} by 2.74 p.p. (Table \ref{tab:robustness_results}). This counter-intuitive finding suggests the "red herring" may have disrupted a brittle heuristic, forcing a more robust reasoning path.  

\begin{table}[h!]  
\centering  
\small  
\begin{tabular}{l|c|c|c|c}  
\toprule  
\begin{tabular}[c]{@{}l@{}}\textbf{Puzzle}\\ \textbf{Type}\end{tabular} & \textbf{Condition} & \textbf{Old} & \textbf{New} & \textbf{Change} \\  
\midrule  
SA & Shuffled & 17.66 & 13.97 & \textcolor{red}{-3.69} \\  
BR & Shuffled & 42.47 & 35.77 & \textcolor{red}{-6.70} \\  
\midrule  
SA & Noisy & 17.66 & 14.58 & \textcolor{red}{-3.08} \\  
BR & Noisy & 42.47 & 45.21 & \textcolor{green}{+2.74} \\  
CD & Noisy & 27.64 & 23.77 & \textcolor{red}{-3.87} \\  
\bottomrule  
\end{tabular}  
\caption{Performance of \textit{Qwen QwQ 32B} under shuffled constraints and irrelevant information. Performance change is in percentage points (p.p.).}  
\label{tab:robustness_results} 
\end{table}  

\section{Conclusion}
\label{sec:conclusion}

In this work, we introduce \textbf{RiddleBench}, a benchmark designed to test complex, multi-step reasoning in LLMs, revealing that even top models struggle significantly. Our analysis uncovers several critical failures in the specific models tested: a \textbf{hallucination cascade}, where the tested models uncritically adopt flawed reasoning of other/evaluated LLMs; strong \textbf{error fixation} preventing the reversal of incorrect judgments; and poor self-correction due to a powerful \textbf{self-confirmation bias}. We also find that LLM reasoning is fragile, easily disrupted by reordered constraints or irrelevant information.

RiddleBench provides the research community with a tool to diagnose these issues and measure progress towards more dependable AI. We plan to expand this effort through further dataset growth and cross-lingual evaluations, as outlined in Section \ref{sec:future_work}.

\section{Limitations}
Our evaluation of proprietary models uses August 2025 API versions; performance may change with updates. RiddleBench’s curation from Indian exam materials may introduce cultural bias, although the logical tasks are universal. High API costs and computation limited cross-model and self-correction experiments to a subset of models.

A key concern is data contamination. We mitigated this by sourcing puzzles from mock exam PDFs, which are rarely part of standard web scrapes. The multi-step deductive nature resists memorization, but contamination cannot be definitively ruled out for proprietary models with undisclosed training sets.

\section{Future Work}
\label{sec:future_work}

While RiddleBench provides a strong foundation for evaluating logical reasoning, we plan to extend this work in several key directions to further probe the capabilities and limitations of LLMs.

First, we intend to expand the benchmark itself. We may increase the number of problems within the existing categories to enhance statistical robustness and potentially introduce new types of logical puzzles to assess a broader spectrum of reasoning skills.

A primary goal for future work is to extend RiddleBench beyond a single language. True reasoning capabilities should be language-agnostic, and evaluating models on multilingual data is critical. We plan to translate the benchmark into several other languages. Given the origin of our source material, we will initially focus on major Indian languages, with the goal of eventually including other world languages. This will enable a more comprehensive, cross-lingual assessment of LLM reasoning and help drive the development of more globally competent models.

\section{Ethics Statement}
Through this work, our aim is to advance the study of reasoning in Large Language Models by creating a challenging and publicly accessible benchmark. By highlighting critical vulnerabilities such as hallucination cascades, poor self-correction, and fragile logic, we hope to guide the development of more robust, reliable, and safer AI systems.

The RiddleBench dataset created in this work is released under the permissible CC0 license to ensure broad and unrestricted access for the research community. Generative AI systems were used only for assistance with language refinement (e.g., paraphrasing, polishing the authors' original content) and for writing boilerplate code.

\section{Acknowledgements}
We would like to thank EkStep Foundation and Nilekani Philanthropies for their generous grant towards research at AI4Bharat.

\bibliography{custom}
\bibliographystyle{acl_natbib}

\appendix
\section{Models Evaluated}
\label{sec:appendix_models}
We evaluated a comprehensive suite of LLMs, encompassing both leading proprietary systems and prominent open-weight models to benchmark their performance.

\subsubsection*{Proprietary Models}
\begin{itemize}
\item \textbf{o3} (OpenAI) \cite{openai2025o3}
\item \textbf{Gemini 2.5 Pro} (Google) \cite{team2023gemini}
\item \textbf{Claude 4 Sonnet} (Anthropic) \cite{anthropic2025claude4}
\end{itemize}

\subsubsection*{Open-Weight Models}
\begin{itemize}
\item \textbf{DeepSeek-R1} (DeepSeek AI) \cite{deepseek2025r1}
\item \textbf{GPT-oss-120B} (OpenAI) \cite{openai2025gptoss120b}
\item \textbf{Mistral Small 24B it} (Mistral AI) \cite{mistral2025small}
\item \textbf{Llama 3.3 70B} (Meta) \cite{meta2024llama3_3}
\item \textbf{Gemma 3 27B it} (Google) \cite{google2025gemma3}
\item \textbf{Qwen QwQ 32B} (Alibaba) \cite{qwen2025qwq32b}
\item \textbf{DeepSeek-V3} (DeepSeek AI) \cite{deepseek2025v3}
\end{itemize}
All experiments were conducted using the latest model versions available as of August 2025 and on API keys.

\section{Dataset Examples}
\label{sec:examples}

\subsection{Coding-Decoding}
\begin{verbatim}
Question: In a certain code 
language FRAME is written 
as QEBDL and BLOCK is 
written as KAPJB. How is 
PRIDE written in that 
code language? 
a) SQHFE 
b) QSHEF 
c) QQJCD 
d) QOJDC 
e) None of these
\end{verbatim}

\subsection{Sequential Reasoning}
\begin{verbatim}
Question: You are given a 
numerical sequence in which 
one term is missing, represented 
by a '?'. Your task is to 
analyze the pattern followed 
by the numbers and determine 
the missing value.
43 41 44 39 46 ?
\end{verbatim}

\subsection{Seating Arrangement}
\begin{verbatim}
Question: Nine persons Anu, Bablu, 
Cheenu, Dona, Esha, Faria, Gaurav, 
Harish and Ishita are sitting 
in a row and all are facing 
north. It is known that Cheenu 
sits exactly in the middle and 
there is no person to the right 
of Ishita. Dona is fourth to 
the right of Faria. Gaurav and 
Harish are sitting next to each 
other. Esha is the neighbor of 
Dona but not of Cheenu. Harish 
doesn't sit at any extreme 
corner. Dona is not sitting 
adjacent to either Cheenu or 
Ishita. Anu is second to the 
right of Harish. Who is sitting 
at the left most seat of the row? 
A. Faria 
B. Bablu 
C. Gaurav 
D. Dona 
E. None of these
\end{verbatim}

\subsection{Blood Relations}
\begin{verbatim}
Question: Soni is brother of 
Moni, Daya is sister of Moni, 
and Bala is father of Charu, 
who is brother of Daya. If Moni 
is son of Roop then how is Bala 
related to Roop?
A. Wife
B. Husband
C. Son
D. Mother in law
E. Brother in law
\end{verbatim}

\section{Prompting Details and Samples}
\label{sec:appendix_prompts}

\subsection{Zero-Shot Evaluation Prompt}
The following prompt was used 
for the main evaluation on 
RiddleBench.
\begin{verbatim}
Question: <question><options if any>.
Please always write the 
final answer in \boxed{}.
Answer:
\end{verbatim}

\subsection{Prompt for Forced-Choice Answer Verification}
\label{subsec:prompt_exp1}

\begin{verbatim}
You are checking a logic based 
sum. For a given question, determine 
out of answer 1 and answer 2 
which is correct. Since the 
answers are for the same question, 
you can assume similar context 
for both answers and make 
appropriate assumptions when 
checking if they are correct. 
Think about both the answers 
and find out which one is correct. 
Please put the final answer as 
"1" or "2" in \boxed{}.

<question>
{question}
</question>

Answer 1: {correct_answer}
Answer 2: {predicted_answer}
\end{verbatim}

\subsection{Prompt for Reasoning Validation}
\label{subsec:prompt_exp2}
\begin{verbatim}
You are a checker for logical 
reasoning questions. You will be 
given a Question, a Reasoning, 
and an Answer. Your task:
(1) Check if the reasoning is 
logically correct.
(2) Check if the answer is correct.
At the end, reply in the format: 
\boxed{YES/NO, The correct answer 
is ...}. Use YES if the reasoning 
and answer are correct, otherwise 
use NO and provide the correct answer.

Question : {question}
Reasoning : {reasoning}
Answer : {answer}
\end{verbatim}

\subsection{Sample Problems for Robustness Experiments}
The following examples illustrate 
the perturbations applied.
\paragraph{Constraint Shuffling}
\begin{verbatim}
Original: A is the father of B. 
B is the sister of C. C is the 
son of D. How is A related to D?

Shuffled: C is the son of D. 
A is the father of B. B is the 
sister of C. How is A related to D?
\end{verbatim}
\begin{figure}[h!]
\centering
\includegraphics[width=0.9\linewidth]{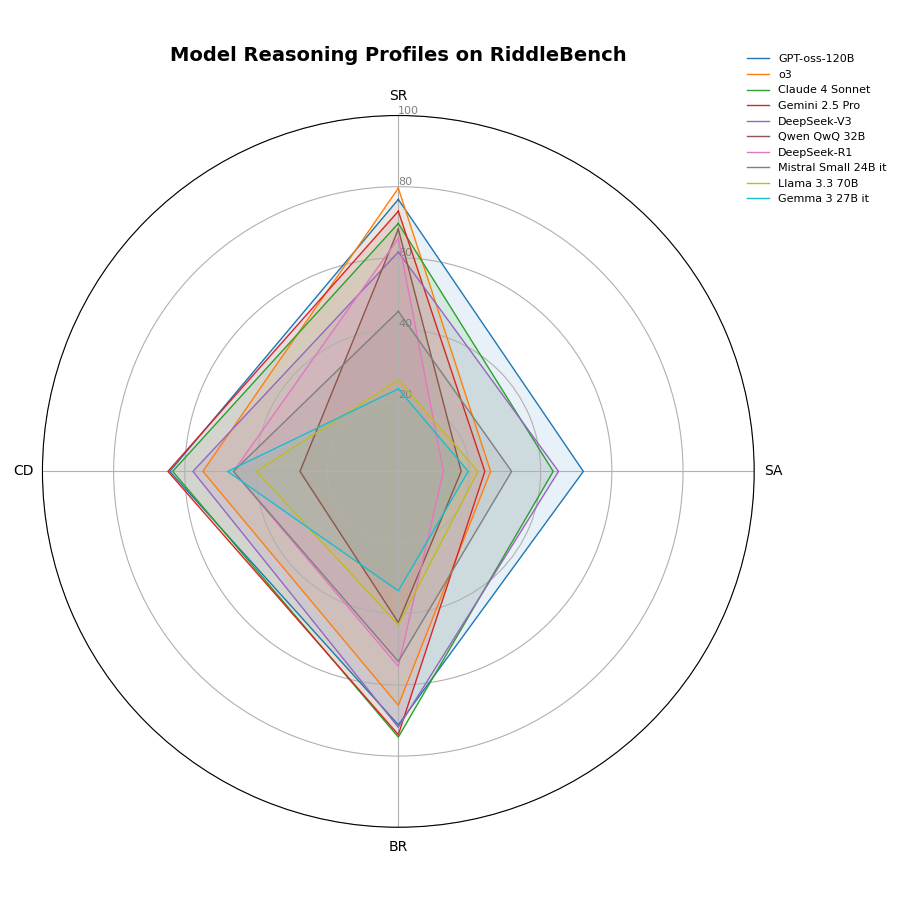}
\caption{A radar chart illustrating the performance of evaluated LLMs across the four reasoning categories of RiddleBench. Each colored line represents a different model, showing its strengths and weaknesses in SR, SA, BR, and CD.}
\label{fig:radar_chart}
\end{figure}
\begin{figure*}[h!]
\centering
\includegraphics[width=1.0\linewidth]{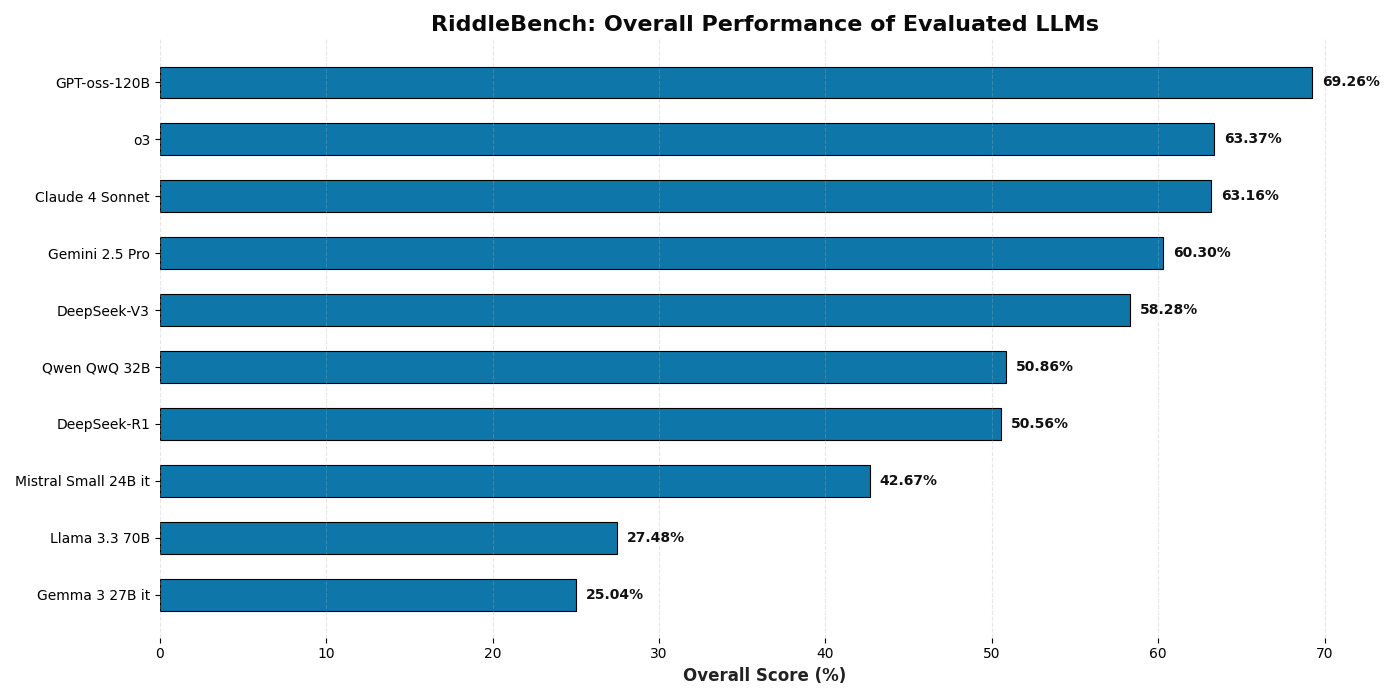}
\caption{Overall performance of evaluated LLMs on RiddleBench. The horizontal bars indicate the percentage of correct answers for each model.}
\label{fig:plt}
\end{figure*}
\paragraph{Irrelevant Information}
The irrelevant sentence is highlighted 
in italics for clarity.
\begin{verbatim}
In a certain code language FRAME 
is written as QEBDL and BLOCK 
is written as KAPJB. *The programmers 
who designed this language often 
take breaks to play basketball.* 
How is PRIDE written in that code 
language?
a) SQHFE 
b) QSHEF 
c) QQJCD 
d) QOJDC 
e) None of these
\end{verbatim}

\section{Model Reasoning Profiles}
\label{sec:appendix_profiles}

Figure \ref{fig:radar_chart} compares LLM reasoning across the four RiddleBench categories: Sequential Reasoning (SR), Seating Arrangement (SA), Blood Relations (BR), and Coding-Decoding (CD). Each axis shows a category, and distance from the center reflects accuracy.

The visualization reveals performance diversity. For example, \textit{GPT-oss-120B} exhibits a large, balanced shape, indicating strong, well-rounded performance. In contrast, other models display more skewed profiles that reveal specific strengths (e.g., in Coding-Decoding) and weaknesses (e.g., in Seating Arrangements). Figure \ref{fig:plt} shows overall LLM performance on RiddleBench.

\section{Visual Aids in Solving Problems}
\label{sec:qualitative_analysis}

\begin{figure}[h!]
\centering
\includegraphics[width=0.9\linewidth]{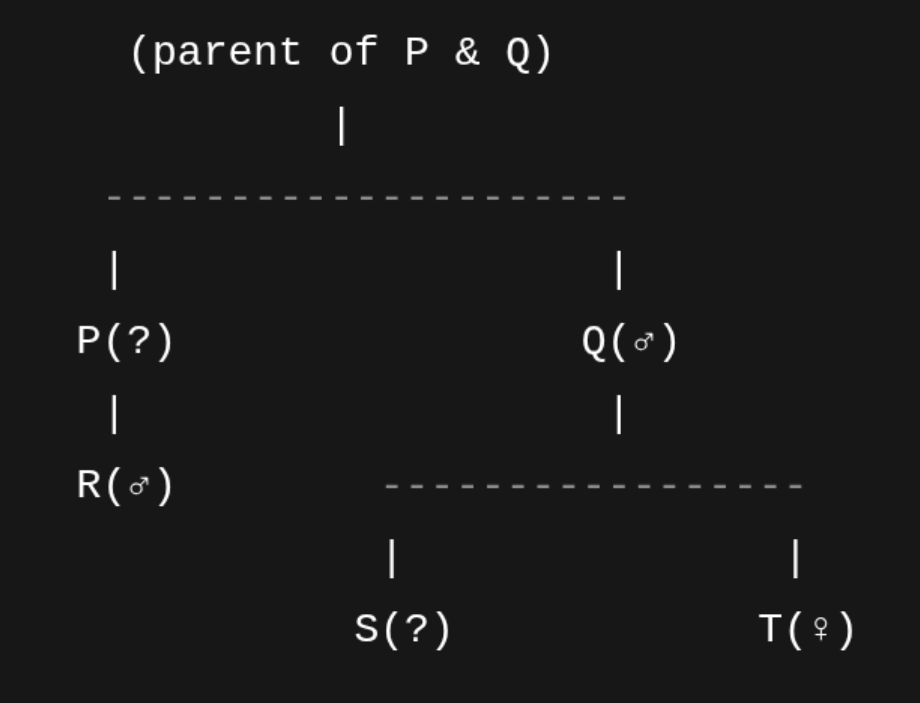}
\caption{An ASCII family tree generated by a Gemini model for a Blood Relations puzzle, a unique visual reasoning strategy observed in our analysis.}
\label{fig:family_tree}
\end{figure}

In our qualitative analysis, we identified a unique reasoning strategy exclusive to Gemini models for solving \textbf{Blood Relations} puzzles: the generation of ASCII art family trees. This emergent behavior, shown in Figure \ref{fig:family_tree}, mimics the human technique of drawing diagrams to visualize complex relationships.

By translating textual constraints into a spatial format, the model attempts a more sophisticated reasoning process than purely sequential text deduction. Although this method does not guarantee a correct answer, the strategy of building a visual mental model represents a promising step towards more robust and interpretable AI reasoning. This approach was not observed in any other model we evaluated.

\section{Patterns in Thinking in Sequence Tasks}
\label{sec:standardized_heuristics}

\begin{figure}[h!]
\centering
\includegraphics[width=0.9\linewidth]{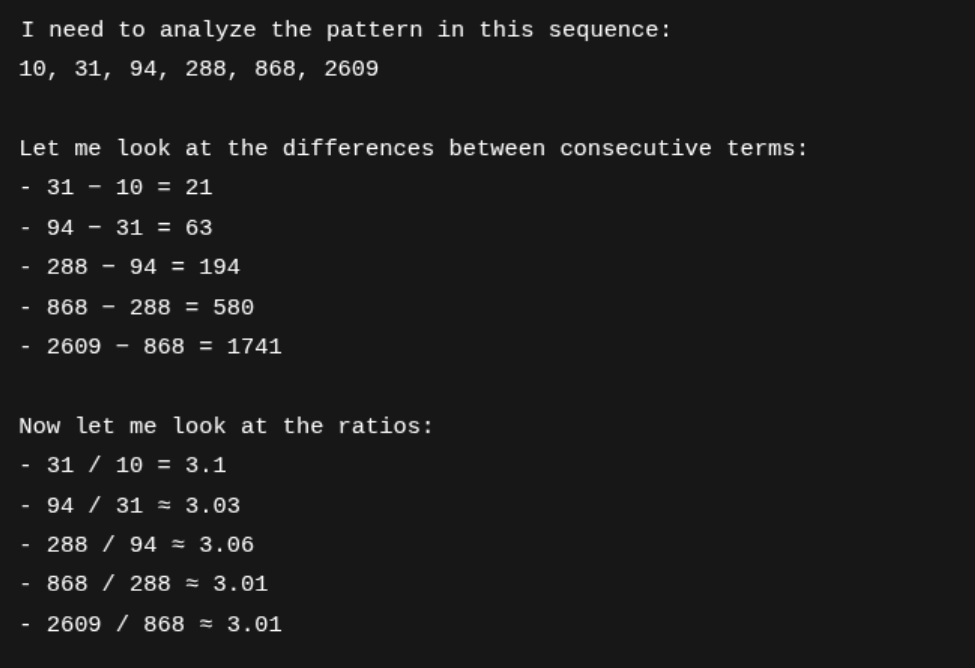}
\caption{A typical reasoning trace from a model solving a numerical sequence. The model first checks for arithmetic differences before successfully identifying a geometric ratio pattern. This two-step heuristic was common across most models.}
\label{fig:sequence_heuristic}
\end{figure}

Our analysis of \textbf{Sequential Reasoning} puzzles revealed that most models use a consistent, two-step heuristic for numerical sequences. As exemplified in Figure \ref{fig:sequence_heuristic}, models first try to find an arithmetic pattern by calculating differences between terms. If that is inconclusive, they pivot to checking for a geometric pattern by calculating their ratios.

This standardized heuristic, which mirrors common human problem-solving methods, suggests the models have learned an effective procedural algorithm. While efficient for typical problems in RiddleBench, this formulaic approach indicates procedural imitation rather than abstract reasoning. This reliance on a fixed template may be a vulnerability for sequences that require more novel or unconventional logic.

\section{Model API Costing}

The models evaluated in this study were accessed through a combination of proprietary APIs and the DeepInfra platform.

The following models were accessed directly via their respective API keys: GPT-oss-120B, o3, Claude 4 Sonnet, and Gemini 2.5 Pro.

All other models were accessed through the DeepInfra API \citep{Deepinfra2025}: DeepSeek-V3, Qwen QwQ 32B, DeepSeek-R1, Mistral Small 24B it, Llama 3.3 70B, and Gemma 3 27B it.

The total cost incurred via API usage for the entire project amounted to \$314. 

\end{document}